\documentclass[letterpaper, 10 pt, conference]{ieeeconf}
\IEEEoverridecommandlockouts
\usepackage{subfigure}
\usepackage{graphicx}
\usepackage{url}
\usepackage{amsmath}
\usepackage{amssymb}
\usepackage{caption}
\usepackage{algorithmic}
\usepackage{algorithm}
\usepackage{amssymb,amsfonts}
\usepackage{color}
\usepackage{float}
\usepackage{listings}
\usepackage[utf8]{inputenc}
\usepackage[english]{babel}
\usepackage[normalem]{ulem}
\usepackage{epstopdf}
\usepackage{booktabs}
\usepackage{cite}
\usepackage{rotating}
\usepackage[colorlinks=true, 
			pdfstartview=FitV, 
			linkcolor=black, 
			citecolor=black, 
			plainpages=false, 
			pdfpagelabels=false, 
			urlcolor=black]{hyperref}
\usepackage[all]{hypcap}
\usepackage{multirow}
\usepackage{xcolor}
\usepackage{colortbl}

\newcommand{\cancel}[1]{}

\title{\LARGE \bf
Motion Planning for Autonomous Vehicles in the Presence of Uncertainty Using Reinforcement Learning 
}

\author{Kasra~Rezaee$^1$, Peyman~Yadmellat$^1$, Simon Chamorro$^{\dagger 2}$ \\
\thanks{$^1$Noah's Ark Lab., Huawei Technologies, Markham, Ontario, Canada, \protect\url{{kasra.rezaee, peyman.yadmellat}@huawei.com}.}
\thanks{$^2$ECE Department, Université de Sherbrooke, Sherbrooke, Quebec, Canada, \protect\url{simon.chamorro@usherbrooke.ca}.}
\thanks{$^\dagger$The work was done during the author's internship at Noah's Ark Lab., Huawei Technologies Canada.}
}

\begin{document}

\maketitle
\thispagestyle{empty}
\pagestyle{empty}

\begin{abstract}
Motion planning under uncertainty is one of the main challenges in developing autonomous driving vehicles. In this work, we focus on the uncertainty in sensing and perception, resulted from a limited field of view, occlusions, and sensing range. This problem is often tackled by considering hypothetical hidden objects in occluded areas or beyond the sensing range to guarantee passive safety. However, this may result in conservative planning and expensive computation, particularly when numerous hypothetical objects need to be considered. We propose a reinforcement learning (RL) based solution to manage uncertainty by optimizing for the worst case outcome. This approach is in contrast to traditional RL, where the agents try to maximize the average expected reward. The proposed approach is built on top of the Distributional RL with its policy optimization maximizing the stochastic outcomes' lower bound. This modification can be applied to a range of RL algorithms. As a proof-of-concept, the approach is applied to two different RL algorithms, Soft Actor-Critic and DQN. The approach is evaluated against two challenging scenarios of pedestrians crossing with occlusion and curved roads with a limited field of view. The algorithm is trained and evaluated using the SUMO traffic simulator. The proposed approach yields much better motion planning behavior compared to conventional RL algorithms and behaves comparably to humans driving style.
\end{abstract}

\section{INTRODUCTION}

Motion planning is the task of finding a trajectory for an autonomous vehicle to follow to achieve its higher level goals \cite{paden2016survey}. 
The most critical objective of motion planning is to deliver a safe trajectory, and in the self-driving context, various sources of uncertainty make this objective challenging.
In this work, we focus on the uncertainty in sensing and perception, resulted from a limited field of view, occlusions, and sensing range.
Conventional approaches tackle this uncertainty by considering hypothetical objects in the occluded regions. 
Thus, safety can be guaranteed if the trajectory avoids all these hypothetical objects.
This approach is reasonably effective when the scope is narrow, and the motion planner needs to cover limited situations.
However, developing a motion planner covering every possible situation is a tedious process, if not impractical, particularly in autonomous driving scenarios.
Furthermore, as scenarios get more complex, such approaches result in conservative planning and expensive computation.

\begin{figure}[t]
    \centering
    \includegraphics[width=0.49\columnwidth]{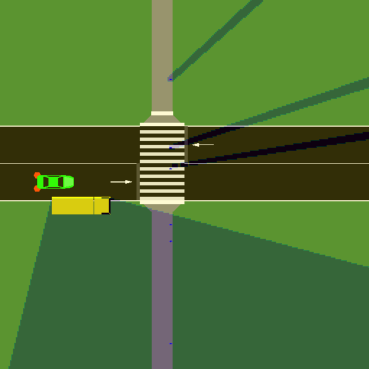}
    \includegraphics[width=0.49\columnwidth, trim={0, 0, 0, 0.15cm}, clip]{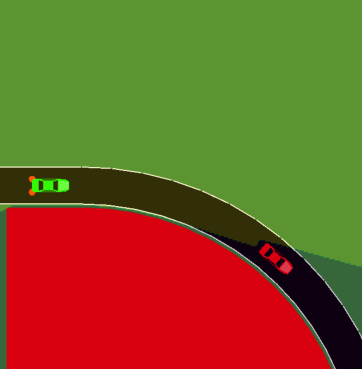}
    \caption{The impact of occlusion on the visibility of an autonomous vehicle and the resulting uncertainty in two
    scenarios. Left, occlusion at the corner of a pedestrian crossing. Right, occlusion as a result of the road curve,
    \textit{e.g.} tunnels or highway off-ramp.}
    \label{fig_sumo_experiments}
\end{figure}

Machine learning and data-driven approaches provide a viable alternative to classic approaches to motion planning in crowded and complex environments. 
However, using machine learning-based approaches without considering the required transparency and safety criteria can be catastrophic in autonomous vehicles.
We aim to address the problem of motion planning in the presence of uncertainty using reinforcement learning (RL). 
Conventional RL formulation aim at maximizing the expected reward, which is not safe and robust to uncertainties. 
We leverage the recent advances in distributional RL, and propose an algorithm that finds the control policy that maximizes performance for the worst-case scenario.
We apply the proposed algorithm to the motion planning problem under occlusion. 
As a proof of concept, we test the algorithm on two driving scenarios: \romannumeral 1) a vehicle passing a pedestrian crossing in the presence of an
occlusion blocking view of approaching pedestrians, and \romannumeral 2) driving on a curved road with the curve blocking the view.
The occlusion is provided to the agent as an occupancy grid map~(OGM) generated from a LIDAR sensor's output. 

To summarize, the main contributions of this paper are: 
\begin{itemize}
\item Adapting the formulation for deep RL problems to take into account uncertainty in the environment and
    optimizes for the worst-case scenario;
\item Developing a motion planning approach that effectively navigates in the presence of occlusion.
\end{itemize}

In Section~II, we review the related works on motion planning. 
In Section~III, the proposed RL-based solution for problems with uncertainty is introduced. 
In Section~IV, the proposed algorithm is utilized to design a motion planning solution. 
Section~V presents a simulation experiment and elaborates the result of applying the proposed algorithm.
Finally, Section~VI summarizes the paper.

\section{RELATED WORKS}

The classic approaches to motion planning usually involve defining a cost function that evaluates trajectories and
searching for the trajectory that optimizes this cost~\cite{paden2016survey}.
Since the trajectories need to satisfy the vehicle's kinematic and dynamic constraints and adhere to social driving
norms, the search space can be reduced to search only within a subset of possible trajectories. 
Classical algorithms model the behavior of various road objects and predict their future position. 
This information is then employed to check the safety of a candidate trajectory.
When dealing with uncertainty due to occlusion, the worst-case possibility (the occluded area being occupied by an object) is considered for safety estimation.
In \cite{8793557}, the ego vehicle's surrounding environment is represented using a grid map, and occluded pixels are propagated in certain directions when predicting future drivable areas.
Considering this constrained drivable area, a trajectory is planned that can guarantee passive safety.
While effective in guaranteeing safety, this assumption leads to excessive computation and conservative planning in complex and crowded environments. 
In \cite{morimura2010parametric}, the authors propose an approach to estimate the state of occluded regions from other users' behavior instead of making a conservative assumption about the occluded regions.
Then the planning is done with the estimated complete state of the environment.

Reinforcement learning provides a solution to deal with the computation cost imposed by crowded environments. 
While in classical approaches, objects need to be explicitly modeled, RL enables estimating the value/cost of a trajectory without modeling individual objects.
There are numerous examples of RL being applied to motion planning \cite{aradi2020survey, kiran2021deep}.
Recent studies of autonomous driving systems using RL have been divided into two main categories: end-to-end approaches and modular systems. 
In the end-to-end approaches, sensory inputs are mapped directly to control commands via trained neural networks~\cite{kendall2019learning, xiao2019multimodal, bojarski2016end}.
The modular systems breake the self-driving problem into smaller problems with each module being developed independently~\cite{feher2019hybrid, ronecker2019deep}.
As such, the motion planner can be developed in isolation.
This hierarchical paradigm allows for better explainability and ease of validation and test.
In our previous work~\cite{rezaee2019multi}, we presented such hierarchical RL planning framework for cruising on multi-lane roads. 
The approach introduced an intermediate abstraction to the motion planning sub-problem, where a behavioral planner dictates a high-level decision that is then carried out by one of many specialized motion planers. 

Conventional RL formulation finds the policy that maximizes the expected sum of future rewards, including the RL methods described previously.
In the presence of uncertainty, unsafe trajectories with catastrophic results rarely happen. 
Therefore, the process of maximizing expected (average) reward does not guarantee the elimination of these unsafe trajectories.
RL formulation can be expanded to maximize objectives that explicitly avoid unsafe events. 
One such objective is conditional value at risk (CVaR). 
In \cite{morimura2010parametric}, a parametric method is employed to estimate the density of the returns. 
Estimating the probability density of the reward allows for optimizing other criteria, other than the expected reward, such as value at risk.
In \cite{chow2018risk, keramati2020being}, RL formulation is modified to maximize CVaR. 
These methods are developed based on solid theoretical grounds and provide proof of convergence and efficiency.
However, they are limited to application with finite states and actions and are not readily applicable to more complex problems.
Quantile networks \cite{dabney2018distributional, pmlr-v80-dabney18a} estimate the reward density with high fidelity without forcing a specific parametric function to the density, which results in a more accurate estimate of the reward density.
In \cite{pmlr-v80-dabney18a}, authors experimented with reward density information to perform optimistic or pessimistic exploration while still maximizing average reward.
Our proposed approach is built on top of the quantile network proposed in \cite{dabney2018distributional} while maximizing a reward associated with the worst-case scenario.

\section{REINFORCEMENT LEARNING TO OPTIMIZE THE WORST CASE}

\subsection{Reinforcement Learning}

The general RL problem is formulated as a discrete Markov Decision Process where an agent interacts with its environment. 
At time step $t$, the agent is in state $s_{t}$ and takes an action $a_{t}$.
The agent obtains a reward $r_{t}$ and transitions to a new state $s_{t + 1}$. 
The goal is to find the optimal policy $\pi^{*}$ that maximize the expected future total reward $R$, 

\begin{equation}
R^\pi = \mathop{\mathbb{E}}\left[\sum\limits_{t=t_0}^{\infty}{\gamma^{t} r(s_{t}, a_{t}) }\right],
\end{equation}
where $\gamma$ is the discount factor.

There are two main categories of RL algorithms: value-based and policy-based.
In the value-based scheme, commonly an action value function $Q(s, a)$ is used to describe the
corresponding state-action pair's value. 

\begin{equation}
Q(s, a) = \mathop{\mathbb{E}_{\pi}}\left[R_{t} | s_{t}=s, a_{t}=a\right]
\end{equation}
This value function is then used to decide the action yielding the best outcome for a given state (the policy). 
On the other hand, in policy-based algorithms, a policy $a_t = \pi(s_t)$ that maps from states to actions is learned directly, with the optimal policy, $\pi^*$, maximizing the expected reward.

\begin{figure}[b]
  \centering
  \includegraphics[width=0.6\columnwidth]{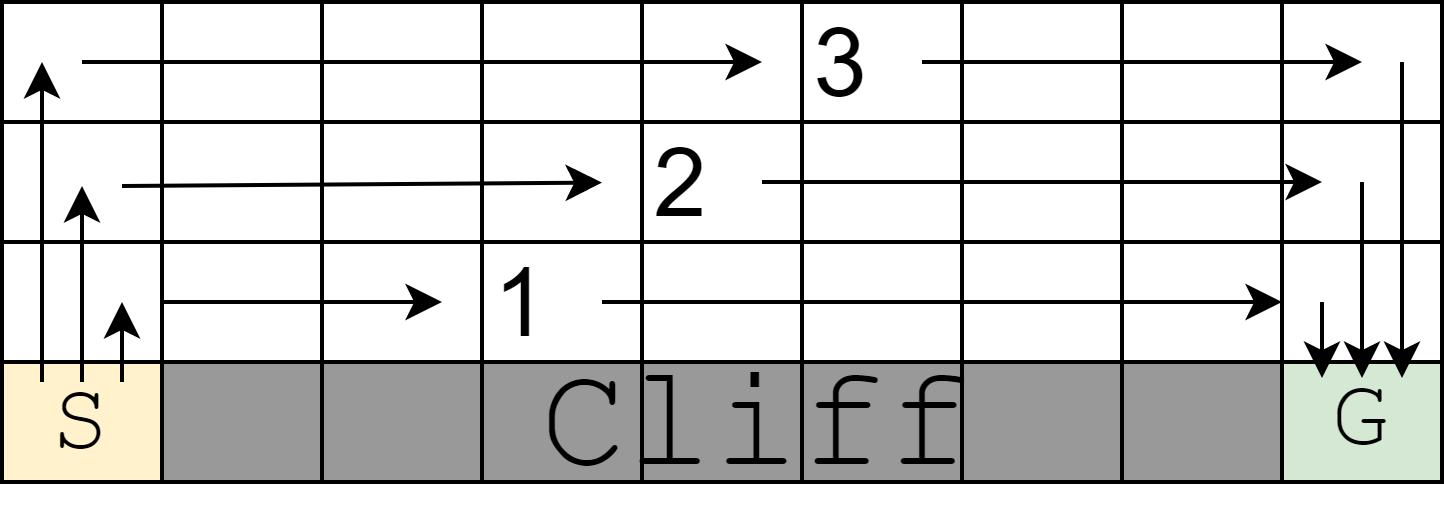}
  \caption{A slightly modified version of the cliff walk example from \cite{sutton1988reinforcement}. 
  The task is to go from start cell S to goal cell G, while avoiding the cliff area.
  The reward is -1 for all states, except the cliff area that has a reward of -20 and terminates the episode.
  Every action is randomly replaced with a down movement with probability $p$.}
  \label{cliff_walk}
\end{figure}

\subsection{The Case Against Maximizing Average Reward}

In RL problems, commonly, the goal is to maximize the expected future total reward.
In problems where the reward is clearly defined (games and toy problems), this approach is ideal as it maximizes the agent performance and provides a level playing field when comparing various algorithms.
However, in practical problems, maximizing the reward is rarely the objective. 
The reward is usually designed to quantify and differentiate various behavior. 
Positive values are assigned for good behavior, and negative values (penalty) are used for behaviors that need to be avoided.

Consider the cliff walk example described in Figure~\ref{cliff_walk}.
Considerin the random down action, moving along path 1 might result in agent falling off the cliff.
For path 2, the agent will have a chance to recover and get away from the cliff if the random down action happens.
For $p=0.1$, the optimal path that maximizes reward would be path 2, as the expected penalty of falling off the cliff outweighs the savings of shorter path 1.
However, for $p=0.01$, the expected penalty is small and optimal path that maximizes the average reward would be path 1.
While the penalty for falling off the cliff can be modified to achieve the desired behavior, in practical applications, many reward values need to be defined, and the probabilities are not known a priori.
This makes it impractical and tedious in real applications to tune the rewards to achieve the desired behavior.
A simple solution to this problem is to assign the worst possible outcome as the value of each state instead of its expected return.
With this defintion for value of a state, path 2 would be optimal irrespective of $p$.

\subsection{Distributional RL to the Rescue}

Distributional RL \cite{dabney2018distributional} aims at estimating the distribution of possible outcomes for each
state-action pair. The total return associated with taking action $a$ in state $s$ and following a policy $\pi$ would be defined by a random variable, $Z^\pi(s,a) = \sum_{t=0}^{\infty} \gamma^t r_t$.
As discussed in the previous section, with access to the distribution of the returns, we can assign the value of a state as the worst case (lower bound) of its possible outcomes: 

\begin{equation}
  Q^\pi (s,a) = \inf (Z^\pi(s,a)),
  \label{equation_lower_bound}
\end{equation}
where $\inf(Z)$ is the lower bound of the random variable $Z$.
One effective approach to estimate the distribution of the random variable $Z$ in the RL context is \textit{Quantile Regression} \cite{dabney2018distributional}.
For a distribution defined with $N$ quantiles $q_1$ to $q_N$, $q_1$ is the approximate lower bound of possible returns.
This approach can be applied to any RL algorithm that incorporates a value function.
For this purpose, the value function needs to be augmented to estimate $N$ quantiles, approximating its distribution.
Then the value of a state would be defined as \eqref{equation_lower_bound}. 

\subsection{Formulation of Conservative Reinforcement Learning}

The QR-DQN \cite{dabney2018distributional} algorithm extends the DQN algorithm to use quantiles as outputs of the network and calculates the value of a state-action
pair as the mean of the distribution defined by the quantiles, $Q^\pi (s,a) = \sum_{j=1}^N{\frac{1}{N} q_j(s,a)}$, where
$q_j(s,a)$ is the value of quantile $j$ for state-action pair $(s,a)$.
When using quantile regression for estimating the value of quantiles, the regression process results in values being sorted from lowest to highest. 
Hence, we can use the first value as the lower bound estimate.
We replace the definition of a state-action pair's value in QR-DQN with 
\begin{equation}
  Q^\pi (s,a) = q_1(s,a),
  \label{eqn:qunatile_low_bound}
\end{equation}
where $q_1$ is the first quantile.
This modification is done when calculating the targets for the learning step and when choosing the optimal action.
We refer to this algorithm as \textit{Conservative QR-DQN} (CQR-DQN)

Another RL algorithm that has attracted attention recently is Soft Actor-Critic (SAC) \cite{haarnoja2018soft} which follows the Actor-Critic framework.
A Q-network is trained to estimate the value of following the policy, and the policy is trained to maximize Q-values.
While in \cite{duan2021distributional}, the authors proposed a distributional extension to the SAC algorithm, they used a Gaussian distribution to represent the random outcomes. 
Since we want to estimate the lower bound, a Gaussian distribution is not suitable for this purpose. 
We propose to extend SAC with quantile regression.
Effectively, the Q-network is extended to estimate the quantiles. Then Q-value of a state-action pair is estimated using~\eqref{eqn:qunatile_low_bound}, similar to the case for QR-DQN. 
Following the distributional Bellman equation in QR-DQN, we can write the distributional SAC Bellman update rule for Critic as:

\begin{equation}
  \label{eqn:dist_bellman}
	Z^{\pi}(s, a)
	:\stackrel{D}{=} 
	r(s, a) + 
	\gamma (Z^{\pi}(s', a') - \log \pi(a' | s')) 
\end{equation}
\noindent
where the sign $:\stackrel{D}{=}$ represent the two sides having the same distribution.
The actor update rule would not change from the original SAC.
We refer to this algorithm as Conservative QR-SAC (CQR-SAC), where QR-SAC refers to the SAC algorithm augmented with quantile regression
while still maximizing the average return.

\section{Motion Planning under Uncertainty}

We aim to solve the motion planning problem when occlusions cause uncertainty.
Consider the case where a vehicle is approaching a pedestrian crosswalk, and a large vehicle parked on the side of the road is blocking the view where pedestrians might enter the crosswalk.
We present this information to the motion planner through an OGM combined with an image of the road network.
The OGM provides information about the occluded areas, and the road network image identifies where road users might be present.
Furthermore, we expect the motion planner to perceive objects from the OGM without providing any explicit information about the objects on the scene.

To solve this motion planning problem, we search for the best trajectory in the Frenet frame.
This is analogous to conventional motion planning in Frenet frame (e.g. \cite{werling2010optimal}).
In Frenet frame, a trajectory that follows the center of the lane becomes a straight trajectory; therefore, simplifying the search space.
We parameterize a trajectory with 4 variables, current speed, $v_0$, current lateral offset, $l_0$, final speed, $v_f$, and final lateral offset, $l_f$.
The trajectory is then created so that the vehicle speed and lateral positions gradually change from initial values to
the final values in a predefined period of time following a first-order exponential trajectory.
While the actual speed is limited to non-negative values, $v_f$ can be negative to help the speed trajectory reach zero faster in emergency situations. 
Additionally, the lateral movement is limited according to the
vehicle's speed. 
The two variables $v_0$ and $l_0$ are based on the vehicle's current state, provided to the motion planner as inputs, and the motion planner needs to optimize $v_f$ and $l_f$.

\subsection{Motion Planning using RL}

The two variables that need to be optimized, $v_f$ and $l_f$, are the actions of the RL agent. 
Inputs to the agent consist of 2 frames of the OGM (current and previous time steps), the current frame of the road
network, and the current speed. 
OGM and road network are 2D images in the vehicle's coordinate frame. 
The current speed is a scalar value that is expanded to fill a 2D input channel.

\begin{figure}[b]
  \centering
  \fbox{\includegraphics[width=0.7\columnwidth{}]{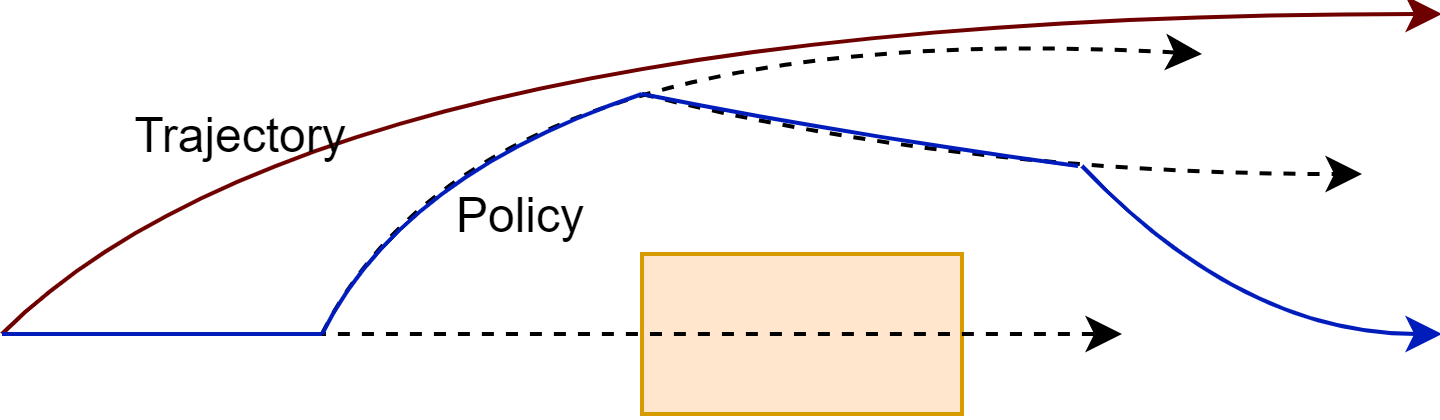}}
  \caption{Illustration of what path would be evaluated when evaluating a trajectory versus a policy.}
  \label{fig_traj_vs_policy}
\end{figure}

We employed a time step of 1 second for transitions between states.
The reward is defined such that it addresses safety, comfort, and mobility needs. 
The mobility reward, $r_m$, is defined to be increasing linearly with the vehicle's speed, $v$ [m/s], up to the speed limit, $v_{lim}$ [m/s], and decreases
quadratically for speeds above the speed limit,
\begin{equation}
  r_m=
  \begin{cases} 
  v,& \text{if } v \leq v_{lim}\\
  \max (0, v - (v-v_{lim})^2),& \text{otherwise}
  \end{cases}.
\end{equation}
The comfort reward, $r_c=-a^2 - |l|$, discourages excessive acceleration and deviation from the center of the lane. 
In this equation, $a$ is the vehicle acceleration [m/$\text{s}^2$], and $l$ is the lateral offset with respect to the lane center [m].
To encourage safety, we terminate the episode and set the reward to zero in case of collision.
To differentiate between a stationary vehicle (also reward of zero) and a collision, the agent receives a reward of $1$ for every time step.
The combined reward of the agent at every step would be:
\begin{equation}
  r=
  \begin{cases} 
  0,& \text{if } \text{collision}\\
  1 + r_m + r_c,& \text{otherwise}
  \end{cases}.
\end{equation}

\subsection{Evaluating a Trajectory Versus Policy}

\begin{figure*}
  \centering
  \includegraphics[height=3.3cm, trim={0mm 7mm 0mm 0mm}, clip]{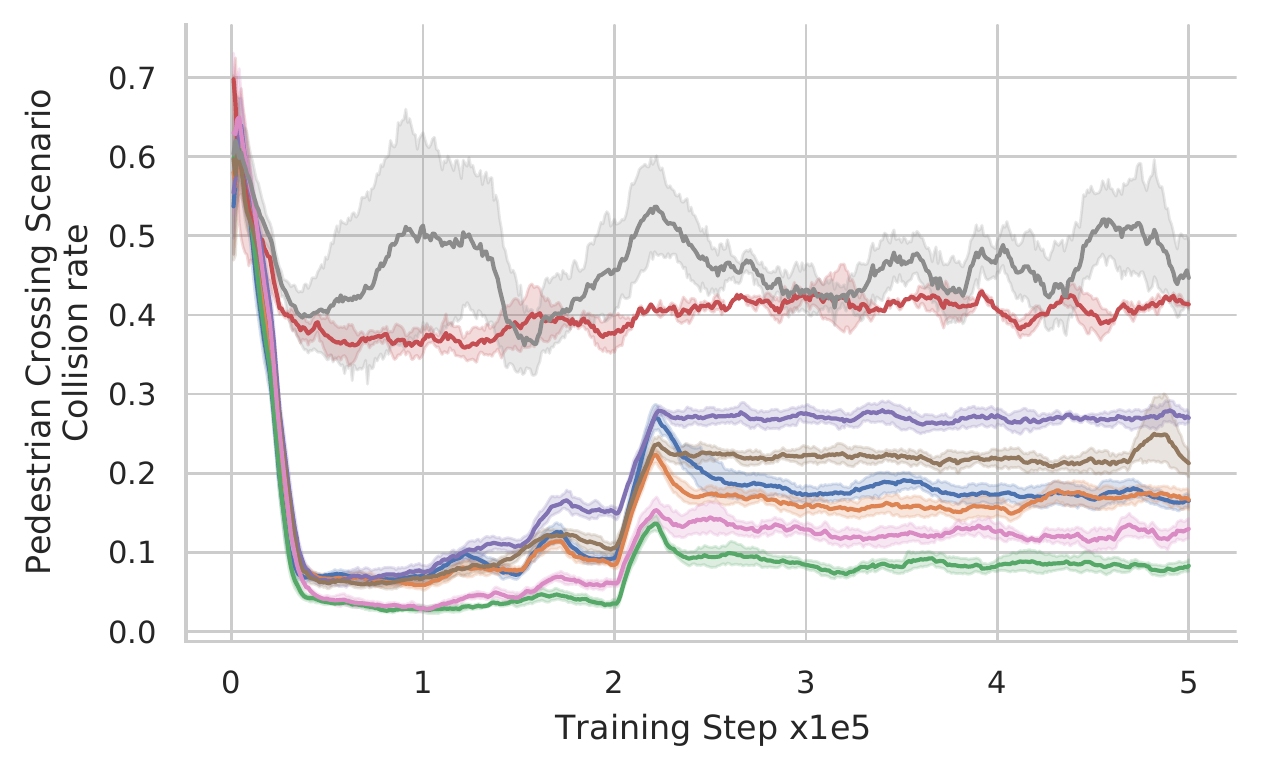}
  \includegraphics[height=3.3cm, trim={0mm 7mm 0mm 0mm}, clip]{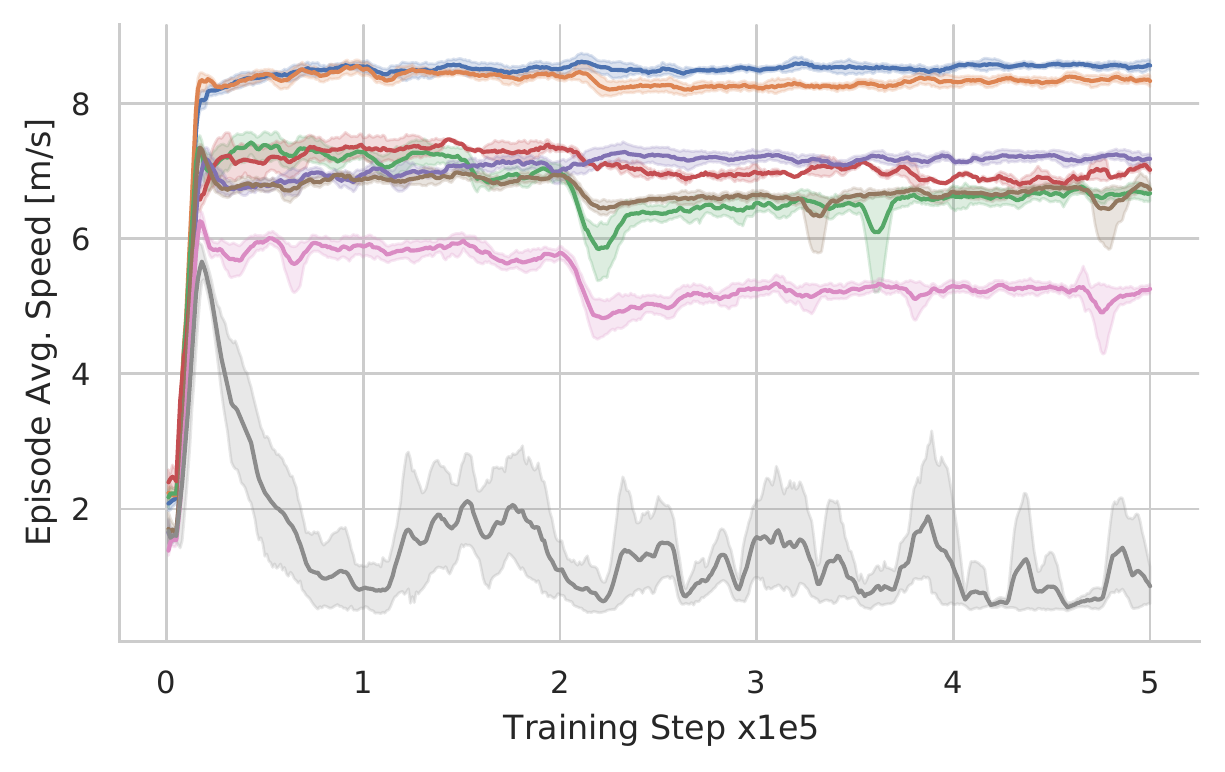}
  \includegraphics[height=3.3cm, trim={0mm 7mm 0mm 0mm}, clip]{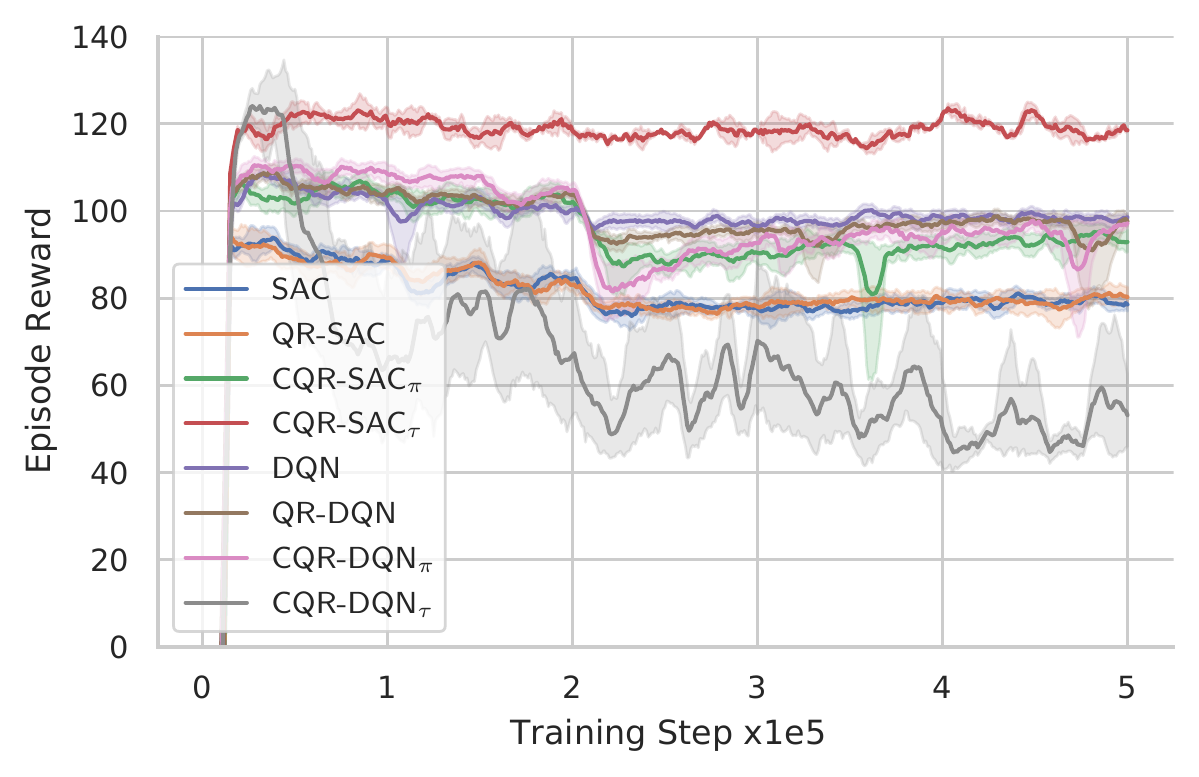}
  \includegraphics[height=3.65cm]{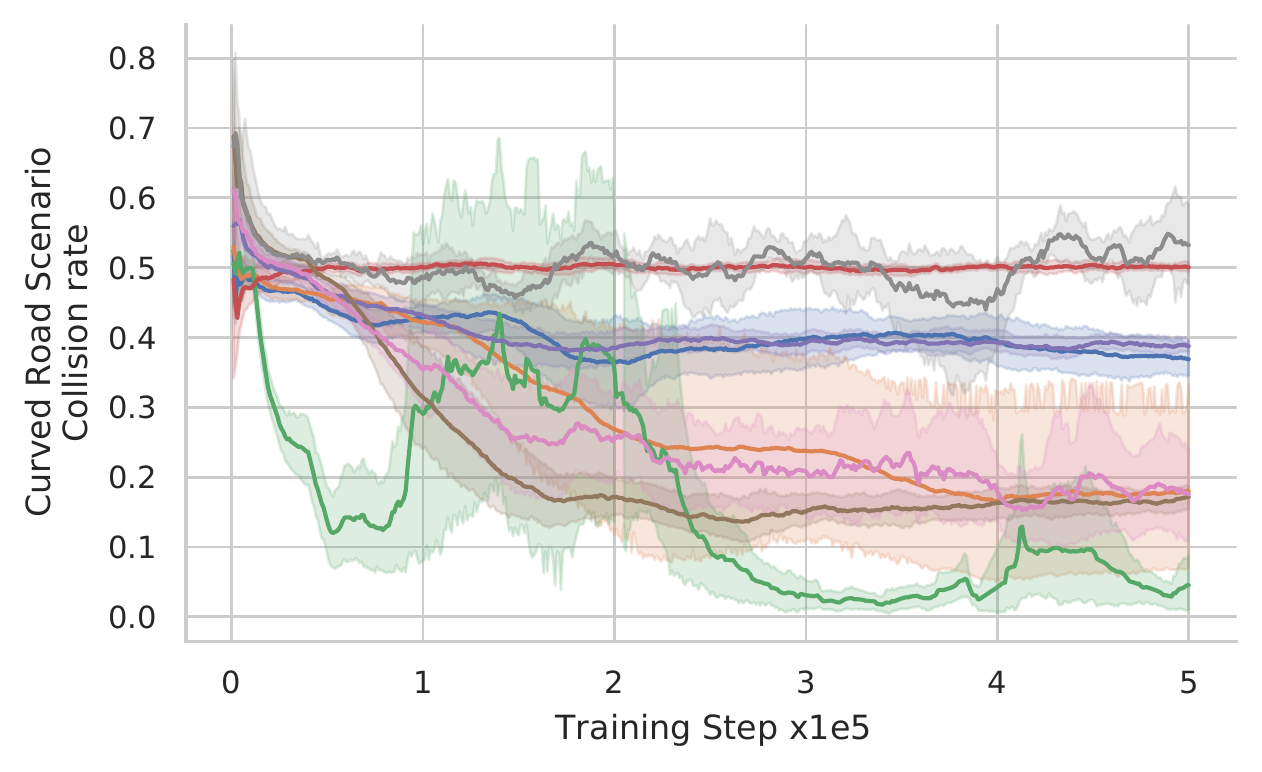}
  \includegraphics[height=3.65cm]{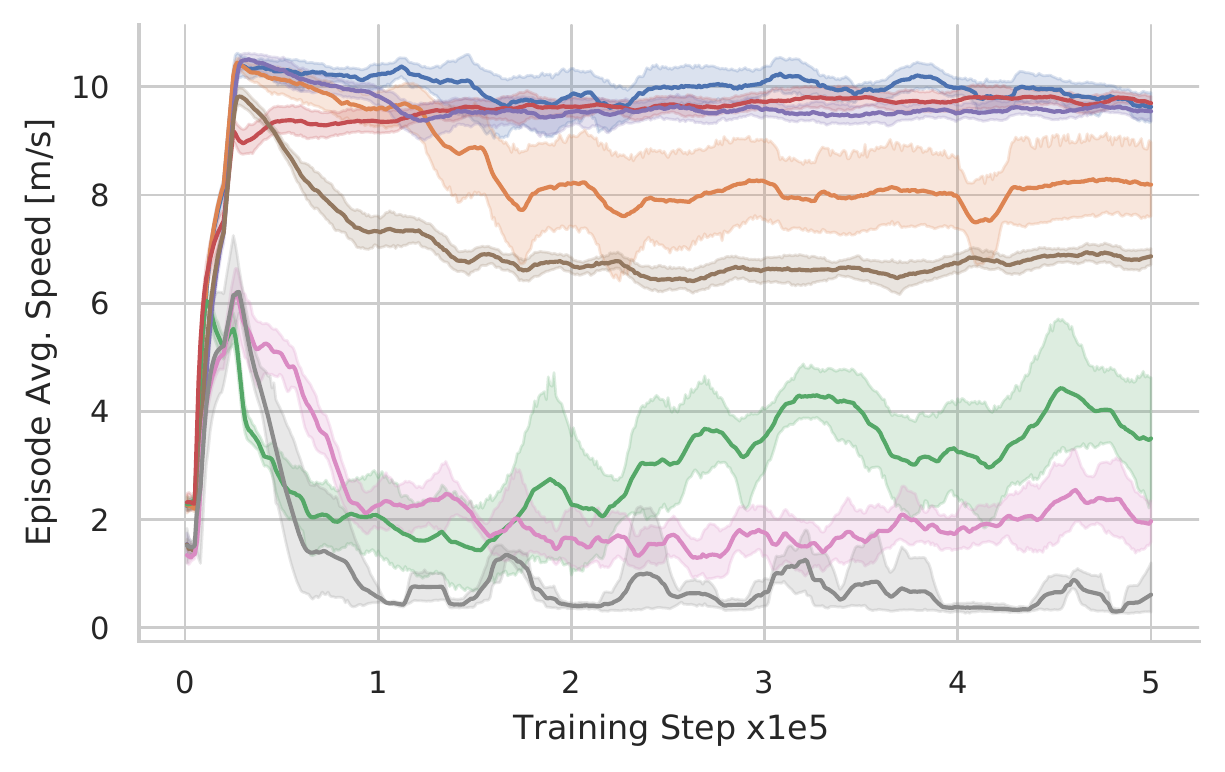}
  \includegraphics[height=3.65cm]{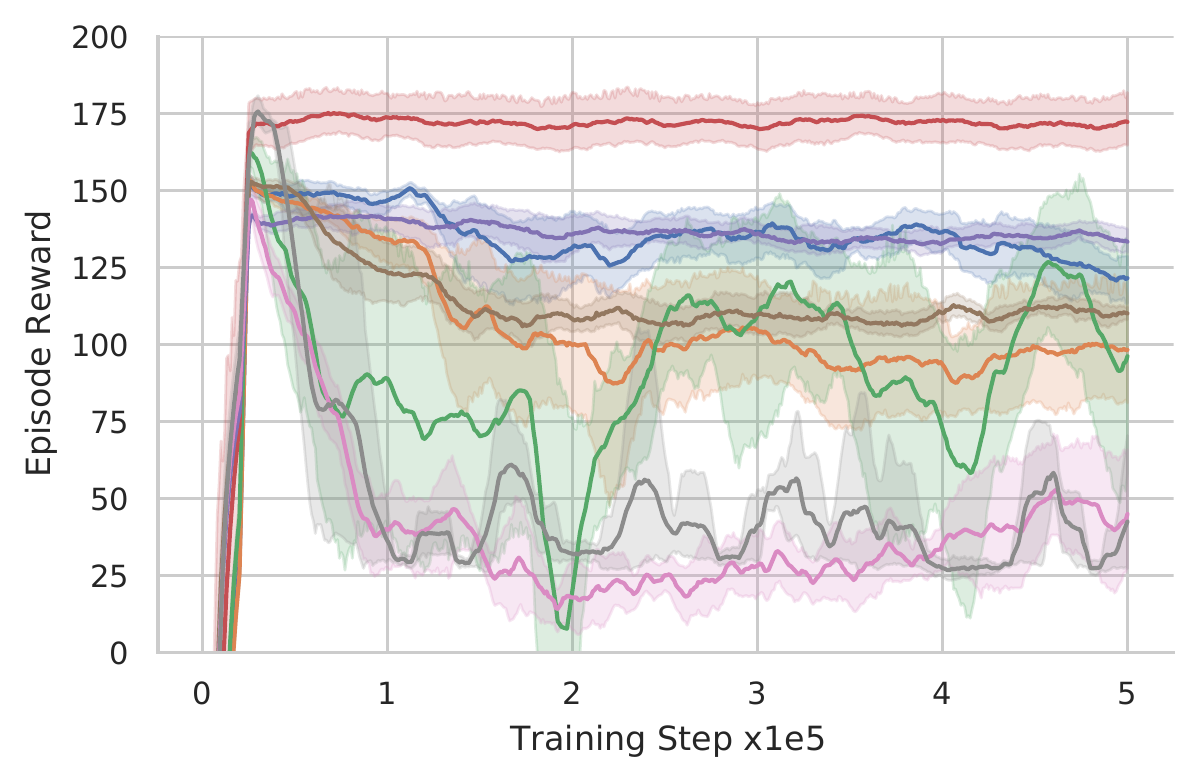}
  \caption{The progress of each algorithm during the training process. 
  The value for each point is the average of the last 1000 training steps. 
  The confidence bands show the variation from 8 training runs with different seeds.
  }
  \label{fig_train}
\end{figure*}

In traditional motion planning, a trajectory is evaluated based on the assumption that the vehicle will follow the
trajectory entirely.
Although the vehicle's trajectory might change in future timesteps, the evaluation will not consider these potential trajectory changes.
From an RL perspective, if the agent's action is defined as the trajectory, evaluating a trajectory is equivalent to
estimate the Q-value with the assumption that the agent's action in future states will be the same as the one in the current state.
The distributional Bellman update rule for such evaluation of a trajectory would be:
\begin{equation}
	Z(s, a)
	:\stackrel{D}{=} 
	r(s, a) + 
	\gamma Z(s', a),
\end{equation}
where $s'$ is the next state, and the current action, $a$, is also used to estimate the next state's value.
We will denote the algorithms that use this update rule with $\tau$ (CQR-DQN$_\tau$, CQR-SAC$_\tau$). 

In contrast, in the RL formulation, the action (trajectory) in future states is according to the agent's policy, and
evaluation is done knowing that the future actions might differ from the current action.
In the case of policy, the Q-value assigned to a state-action pair is the expected reward if the agent's policy is followed. 
Figure~\ref{fig_traj_vs_policy} illustrates the path that would be evaluated when evaluating a trajectory versus a policy.
Following and evaluating a policy will result in more flexibility, and the motion planner can potentially find better solutions.
We will denote the algorithms that evaluate the agent's policy with $\pi$ (CQR-DQN$_\pi$, CQR-SAC$_\pi$).

\section{EXPERIMENTS}
\begin{table*}[]
\caption{Summary of the test performance of the various algorithms on the \textit{curved road} and \textit{pedestrian crossing} scenarios.}
\label{table_evaluation}
\resizebox{\textwidth}{!}{
\begin{tabular}{llrrrr|rrrr|rrr}
\toprule
Scenario                                           & Metric             & SAC    & QR-SAC & CQR-SAC$_\pi$ & CQR-SAC$_\tau$ & DQN    & QR-DQN & CQR-DQN$_\pi$ & CQR-DQN$_\tau$ & Fixed  & Naive  & Aware  \\ \midrule
\multirow{4}{*}{}                                  & $\bar{r}$          & 133.12 & 110.73 & 121.6          & 164.88          & 141.91 & 128.53 & 35.75         & 23.81          & 147.59 & 140.84 & 134.79 \\ 
\rowcolor{gray!20} \cellcolor{white}Curved         & Collision Rate \%  & 37.56  & 26.05  & 3.71           & 49.23           & 38.66  & 20.26  & 14.47         & 38.23          & 50.47  & 26.92  & 0.0    \\ 
Road                                               & $\bar{v}$ m.s$^{-1}$ & 9.90   & 8.36   & 4.15           & 11.01           & 10.52  & 8.04   & 2.26          & 0.94           & 14.54  & 8.81  & 7.26   \\ 
\rowcolor{gray!20} \cellcolor{white}               & a m.s$^{-2}$ (5\textsuperscript{th} \%) & -2.47  & -3.50  & -2.21          & -1.19           & -4.0   & -4.0   & -2.0          & 0.0            & 0.64   & -2.66  & -3.07  \\ \midrule
\multirow{4}{*}{} & $\bar{r}$                      & 75.86  & 83.43  & 103.01         & 109.93          & 102.50 & 105.24 & 104.97        & 21.01          & 114.18 & 116.95 & 100.51 \\ 
\rowcolor{gray!20} \cellcolor{white}Pedestrian     & Collision Rate \%  & 17.19  & 17.81  & 6.38           & 41.96           & 24.29  & 16.7   & 10.1          & -              & 45.31  & 27.25  & 4.03   \\ 
Crossing                                            & $\bar{v}$ m.s$^{-1}$ & 9.25   & 9.11   & 7.03           & 9.81            & 8.13   & 7.46   & 5.68          & 0.0            & 9.51   & 8.28   & 6.04   \\ 
\rowcolor{gray!20} \cellcolor{white}               & a m.s$^{-2}$ (5\textsuperscript{th} \%) & -3.08  & -2.87  & -1.83          & 0.0             & -2.0   & -2.0   & -2.0          & 0.0            & 0.17   & -1.96  & -3.43  \\ \bottomrule
\end{tabular}
}
\end{table*}

We test the proposed algorithms on two scenarios involving occlusion, as shown in Figure~\ref{fig_sumo_experiments}. 
The first task is to drive through a pedestrian crosswalk with an object at the corner of the crossing, blocking the
car's view of the pedestrians approaching the crosswalk.
Pedestrians are randomly spawned and traverse the road at the crossing.
The second task involves a curved road with the side of the road having a large barrier, analogous to real-life examples such as tunnels and off-ramps in urban areas.
For half of the episodes, we place a stationary vehicle at a random position on the road, replicating a traffic jam.
The various agents were trained and tested on scenarios developed using the SUMO traffic simulation software \cite{SUMO2018}.

We used a curriculum approach to gradually increase the task's difficulty from 1 to 5 every 50k steps, with the complexity being 5 from 200k steps onward.
As the complexity increases, the occlusion is moved closer to the road.
The curriculum helps the agent initially learn to move and adopt a more prudent behavior and slow down as occlusion blocks its view.

For each task, we trained and compared the following RL algorithms: SAC, QR-SAC, CQR-SAC$_\pi$, CQR-SAC$_\tau$, DQN, QR-DQN, CQR-DQN$_\pi$, CQR-DQN$_\tau$.
Additionally, as baselines, we developed and tested three rule-based planners: \textit{fixed}, \textit{naive}, and \textit{aware}. 
The \textit{fixed} planner drives at the speed limit and does not consider other objects.
The \textit{naive} planner ignores the occlusion and drives at the speed limit unless it sees an object in its driving path.
In such cases, it will slow down with a constant deceleration, up to a maximum of -4 [m/s$^2$], with the target of stopping just before the object.
The \textit{aware} planner is based on the IADSR algorithm presented in \cite{8793557} and is aware of the occlusion. 
It assumes that an object is present in the occluded area and drvies at a speed that makes it possible to slow down (with -4 [m/s$^2$] deceleration) to a full stop without collision, if an object emerges from the occluded area.
Additionaly, the \textit{aware} planner also maneuvers away from the occlusion to increase its view around the occlusion.

\subsection{Training and Evaluation Result}
Each agent was trained for a total of 500k training steps with 8 different seeds.
Figure~\ref{fig_train} show the training progress of each agent.
Note that performance drops at 50k step intervals, up to 200k, as the problem complexity increases. 
If we just focus on the average episode reward, we might conclude that the CQR-SAC$_\tau$ algorithm had the best performance.
However, its higher collision rates show that it was not successful in learning to avoid collisions.

When focusing on the collision rate, the CQR-SAC$_\pi$ and CQR-DQN$_\pi$ perform better than the rest and inline with the desired behavior.
For the pedestrian crossing scenario, the SAC-based algorithms fared better than DQN-based algorithms.
Also, extensions of SAC and DQN with quantile regression did not help significantly, showing that SAC and DQN algorithms were already successful in estimating the expected reward. 
However, in the curved road scenario, quantile regression significantly improves the collision rate compared to SAC and DQN. 

We evaluated the final trained agents (all 8 seeds for each agent) for 10k steps in test mode without the random action sampling.
Table~\ref{table_evaluation} summarizes the average performance of various agents in addition to the rule-based planners.

\begin{figure}
    \centering
    \includegraphics[width=0.44\textwidth]{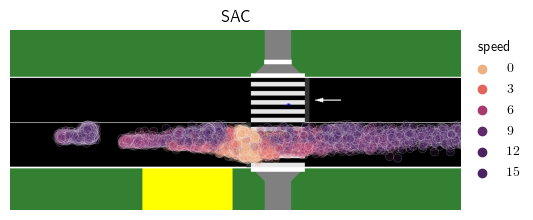}
    \includegraphics[width=0.44\textwidth]{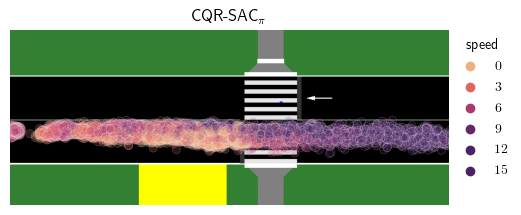}
    \caption{The comparison between behavior of two motions planners, SAC and CQR-DQN$_\pi$. The dots show the vehicles'
    positions across multiple episodes, with the color representing the vehicle's speed.}
    \label{fig_behavior}
\end{figure}

The \textit{fixed} agent gives us the baselines for the worst performance.
As expected, the \textit{naive} agent fails to prevent collisions when the obstacle is in the occlusion areas.
But, when the obstacle is not behind occlusion, it prevents collisions and has lower collisions compared to \textit{fixed} agent.
The \textit{aware} agent represents the highest possible performance in terms of safety and avoids all collisions for the curved scenario.
For the pedestrian crossing scenario, the collisions happen due to the pedestrians' random behaviors and how SUMO simulates their movement.  
The RL-based agents repeat their performance from the training plots, though slightly better as actions are deterministic.

In the pedestrian crossing scenario, we expected that algorithms that maximize average reward (SAC, QR-SAC, DQN, and
QR-DQN) yield higher overall reward; however, the result shows that maximizing the lower bound of the reward can result
in an overall higher reward.
We conclude that this is mostly due to penalty from excessive deceleration based on the agents' behavior analysis.
The agents that approach objects with high speed need to brake more firmly to stop and receive a substantial negative reward.

The vehicle's movements along the road for two motion planners, SAC and CQR-SAC$_\pi$, are shown in Figure~\ref{fig_behavior}.
The color of the dots shows the speed of the vehicle at that point. 
We can see that both planners have the same speed after the intersection, but the CQR-SAC$_\pi$ agent approaches the
crosswalk with much slower speeds compared to the SAC agent.
Furthermore, it is interesting that both RL-based planners have learned that they need to move to the left as they
approach the crosswalk to have a better view behind the occlusion.

\section{CONCLUSIONS \& FUTURE WORK}
In this work, we targeted the motion planning problem in the presence of uncertainty caused by occlusion.
We discussed how in practical RL problems, a policy that maximizes the worst-case reward could better match the desired behavior and leverage the distributional RL to maximize the worst case instead of the average reward.
Extensions to SAC and DQN using quantile regression were proposed to find the action that optimizes the worst-case scenario.
As a proof of concept, a set of motion planners for the self-driving task in the presence of occlusion were designed and evaluated using SUMO simulation environment. 
The results show that in problems where the reward is defined to achieve certain behavior, the reward alone cannot provide a
useful metric for assessing the RL agents' performance.
Furthermore, our proposed motion planners based on CQR-SAC$_\pi$ and CQR-DQN$_\pi$ achieved the desired behavior of
avoiding collision with an occluded view without requiring to fine-tune the reward.

We plan to apply the proposed approach to more complex and diverse environments in our future works.
Such environments include intersections, roundabouts, and the inclusion of moving vehicles in the scenarios. 
With the inclusion of other moving vehicles, our expectation is that the ego agent can implicitly infer the state of the occluded area from other vehicles' behaviors.


\bibliographystyle{IEEEtran}
\bibliography{bibliography} 

\end{document}